\documentclass[letterpaper, 10 pt, journal, twoside]{IEEEtran}  

\IEEEoverridecommandlockouts                              

\usepackage{graphicx}
\graphicspath{{mosaic/}{figures/}}
\usepackage{subcaption}
\usepackage{array}
\usepackage{multirow}
\usepackage{hyperref}
\usepackage{cite}

\title{Failing to Learn: Autonomously Identifying Perception Failures for Self-driving Cars}

\author{Manikandasriram~Srinivasan~Ramanagopal$^{1}$, Cyrus~Anderson$^{1}$, Ram~Vasudevan$^{2}$ and Matthew~Johnson-Roberson$^{3}$
\thanks{Manuscript received: February, 24, 2018; Revised May, 30, 2018; Accepted July, 3, 2018.}
\thanks{This paper was recommended for publication by Editor Jana Kosecka upon evaluation of the Associate Editor and Reviewers' comments. 
This work was supported by a grant from Ford Motor Company via the Ford-UM Alliance under award N022884.}
\thanks{$^1$M. Srinivasan~Ramanagopal and C. Anderson are with Robotics Institute, University of Michigan, Ann Arbor, MI 48109 USA {\tt\footnotesize srmani,andersct@umich.edu}}%
\thanks{$^2$R. Vasudevan is with the Mechanical Engineering, University of Michigan, Ann Arbor, MI 48109 USA {\tt\footnotesize ramv@umich.edu}}%
\thanks{$^3$M. Johnson-Roberson is with the Department of Naval Architecture and Marine Engineering, University of Michigan, Ann Arbor, MI 48109 USA {\tt\footnotesize mattjr@umich.edu}}%
\thanks{Digital Object Identifier (DOI): 10.1109/LRA.2018.2857402}
}

\begin{document}
\maketitle

\IEEEpubid{\begin{minipage}{\textwidth}\ \\[45pt]
\centering
\copyright~2018 IEEE. Personal use of this material is permitted. \\
Permission from IEEE must be obtained for all other uses, in any current or future
media, including reprinting/republishing this material for advertising or
promotional purposes, creating new collective works, for resale or
redistribution to servers or lists, or reuse of any copyrighted
component of this work in other works.
\end{minipage}}

\begin{abstract}
One of the major open challenges in self-driving cars is the ability to detect cars and pedestrians to safely navigate in the world. 
Deep learning-based object detector approaches have enabled great advances in using camera imagery
to detect and classify objects. 
But for a safety critical application such as autonomous driving, the error rates of the current state-of-the-art are still too high to enable safe operation. 
Moreover, the characterization of object detector performance is primarily limited to testing on prerecorded datasets. 
Errors that occur on novel data go undetected without additional human labels. 
In this paper, we propose an automated method to identify mistakes made by object detectors \emph{without ground truth labels}. 
We show that inconsistencies in object detector output between a pair of similar images can be used as hypotheses for false negatives (e.g. missed detections)
and using a novel set of features for each hypothesis, an off-the-shelf binary classifier can be used to find valid errors.
In particular, we study two distinct cues - \emph{temporal} and \emph{stereo} inconsistencies - using data that is readily available on most autonomous vehicles.
Our method can be used with any camera-based object detector and we illustrate the technique on several sets of real world data. 
We show that a state-of-the-art detector, tracker and our classifier trained only on synthetic data can identify valid errors on KITTI tracking dataset with 
an Average Precision of $0.94$. We also release a new tracking dataset with $104$ sequences totaling $80,655$ labeled pairs of stereo images along with ground
truth disparity from a game engine to facilitate further research. The dataset and code are available at \url{https://fcav.engin.umich.edu/research/failing-to-learn}
\end{abstract}


\section{INTRODUCTION}
\IEEEPARstart{O}{bject} detection in self-driving cars is one of the most challenging and important impediments to full autonomy. Self-driving cars need to be able to detect cars and pedestrians to safely navigate
their environment. In recent years, state-of-the-art deep learning approaches such as 
Convolutional Neural Networks (CNNs) have enabled great advances in using camera imagery
to detect and classify objects. In part these advances have been driven by benchmark datasets that have large amounts of labeled training data (one such example is the KITTI~\cite{geiger2013vision} dataset). Our understanding of how well we solve the object detection task has largely been measured by assessing how well novel detectors perform on this prerecorded human labeled data. Alternatives such as simulation have been proposed to address the lack of extensive labeled data~\cite{johnson2017driving}. However, such solutions do not directly address how to find errors in streams of novel data logged from fleets of deployed autonomous vehicles (AVs).

The current AV testing pipeline of repeatedly gathering and labeling large test datasets to 
benchmark object detector success is time-consuming and arduous. This solution does not scale well as object detectors fail less frequently and as the number of deployed AVs increase. A typical AV has multiple cameras, each camera capturing tens of images per second, and hundreds of such AVs could be deployed in a city; all of these images are being processed by the object detector algorithm which could potentially miss objects in any of these images. 
Without hand labeled ground truth, understanding when an object detector has failed to recognize an object is a relatively unstudied problem for self-driving cars. 

This paper introduces a novel automated method to identify mistakes made by
object detectors on the raw unlabeled perception data streams from an AV. The proposed system allows AVs to continuously evaluate their object detection performance in the real world for different locations, changing weather conditions
and even across large time scales when the locations themselves evolve. As testing groups of AVs becomes more commonplace this approach provides an unsupervised mechanism to understand algorithmic, spatial, temporal, and environmental failures of a system's perception stack at the fleet level.

Detecting mistakes within unlabeled data is an inherently ill-posed problem. Without relying on additional data it is fair to assume an object detector will be maximizing its use of the information contained within a single image. We leverage the inherent spatial and temporal nature of the AV object detection domain. Typically either the vehicle or the objects in the scene are moving and often multiple views of the scene are taken (often with overlap as in the case of stereo cameras). While it is important to note object trackers also utilize temporal information (and in fact we leverage trackers for our approach) we are not attempting to solve the tracking problem. Object \textit{detectors} are still required on AVs to initialize a tracker and if an object detector fails to fire, the tracking system is of no use and accidents may ensue. For example if an object detector fails to identify a car as it pulls out of a driveway, finding it 10 frames later and tracking it from that point may be moot as a collision may have already occurred.

We propose that inconsistencies in object detector output between a pair of similar
images (either spatially or temporally), if properly filtered, can be used to identify errors as a vehicle traverses the world. The power of this should not be understated. It means that even miles driven by humans for testing purposes can be used to validate object detectors in an unsupervised manner and furthermore any archives of logged sensor data can be mined for the purposes of evaluating a vehicle's perception system.

The key contributions of our paper are as follows: 1) We present the first full system, to the best of our knowledge, that autonomously detects errors made by single frame object detectors on unlabeled data; 2) We show that inconsistencies in object detector output between pairs of similar images - spatially or temporally - provides a strong cue for identifying missed detections; 3) We pose the error detection problem as binary classification problem where for each inconsistent detection, we propose novel set of meta classification features that are used to predict the likelihood of the inconsistency being a real error; 4) In conjunction with additional localization data available in AV systems we show that our system facilitates the analysis of correlations to geo-locations in errors; 5) We release a tracking dataset with sequences of stereo images gathered at $10$ Hz with ground truth labels following the KITTI format with $104$ sequences totaling $80,655$ pairs of images along with ground truth disparity maps from a game engine making it the largest publicly available dataset of its kind. 

The remainder of the paper is structured as follows. In Section~\ref{sec:related_work}, we discuss related work. Next, we detail our technical approach in Section~\ref{sec:methods}. Section~\ref{sec:result} contains extensive experimental results on a number of datasets for different state-of-the-art object detectors followed by a discussion in Section~\ref{sec:discussion}. Finally, Section~\ref{sec:conclusion} concludes and addresses future work.


\section{Related work}
\label{sec:related_work}

Over the years, researchers have regularly analyzed the performance of object detectors on labeled datasets to understand causes and correlations in errors \cite{dollar2012pedestrian, agrawal2014analyzing, zhang2016far}. Most recently, Zhang et. al.~\cite{zhang2016far} begin by creating a human baseline for the Caltech Pedestrian dataset and use this to aid their analysis of state-of-the-art CNN based detectors. Their analysis is dependant on the extensive manual annotation of a dataset, removing the human in this loop is the exact problem we attempt to solve. In contrast, the methods presented in these papers could be applied to understand the errors identified by our proposed system.

Introspective classification~\cite{grimmett2016introspective,blair2014introspective} has been proposed in the context of classification systems employed in safety-critical applications where errors have severe consequences. In contrast to the commonly used precision-recall metrics, these works introduce introspective capacity as a metric for choosing classification algorithms. In particular, they argue that when presented with an unusual test datum, classification algorithms must respond with high uncertainty. 
Similarly, recent works have focused on estimating the difficulty of a given input image or location in order to make informed decisions. 
Zhang et al.~\cite{zhang2014predicting} propose the ALERT framework which assesses the difficulty of a given input image irrespective of the specific algorithm used for a vision task. More recently, Daftry et. al.~\cite{daftry2016introspective} apply a similar idea by using spatio-temporal features from a deep network. In a similar fashion, Gur\u{a}u et. al.~\cite{guruau2017learn} build a place specific performance record of a vision system which facilitates self-evaluation. 
 However, these methods cannot be directly used to flag missed objects. In contrast, our system would benefit from reliable confidence estimates as this would reduce false positives of object detectors. 

In a similar spirit to our proposed approach, differences in output from pairs of mirror images \cite{yang2015mirror} or from multiple models processing the same image \cite{pei2017deepxplore} have been used to evaluate vision systems.  In \cite{yang2015mirror}, mirrorability is proposed as a measure for evaluating performance of vision systems. However, a typical ensemble detector could consider mirroring as one of the test time data augmentation techniques amongst others such as cropping and shifting. In \cite{pei2017deepxplore}, multiple deep neural networks addressing a common task are cross-referenced and the inconsistencies are used to flag errors of individual networks. This is in principle similar to using an ensemble of single frame object detectors where the ensemble is expected to be a better object detector than individual components. Such ensemble detectors, either with test time data augmentation or multiple separate models trained for the same task, could easily replace the individual detectors in our proposed system and such work compliments the proposed approach. As the ensemble detector is still single image-based it is not using the spatially or temporal information exploited here. 

Sensor fusion poses a related problem by using multiple modalities to understand the world and in some cases identify errors or provide self-supervised labeling~\cite{barnes2017find}. However, in this work we focus on identifying errors within camera imagery alone. This enables single frame object detectors to be independently validated and improved without the confounds of other modalities. 


\section{Technical Approach}
\label{sec:methods}

\begin{figure*}[t]
    \centering
    \begin{subfigure}[t]{\linewidth}
        \includegraphics[width=\textwidth]{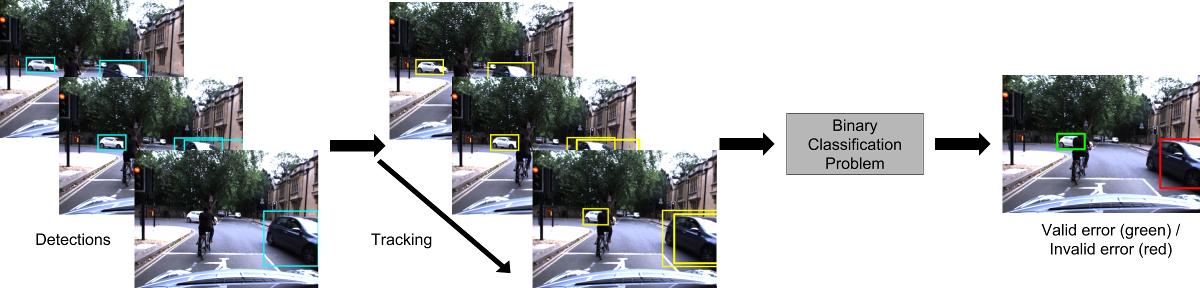}
        \caption{Pipeline for Temporal Cue}
        \label{fig:pipeline_temporal}
    \end{subfigure}
    \begin{subfigure}[t]{\linewidth}
        \includegraphics[width=\linewidth]{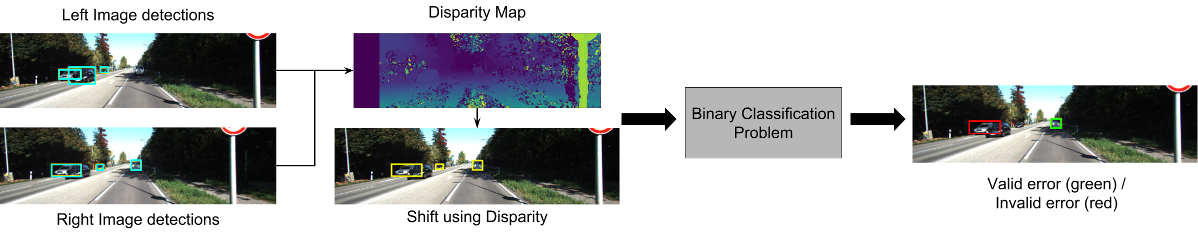}
        \caption{Pipeline for Stereo Cue}
        \label{fig:pipeline_stereo}
    \end{subfigure}
    \caption{(a) Images from the Oxford RobotCar dataset~\cite{maddern20171} are shown with detections (cyan), tracks (yellow), errors predicted by classifier (green) and non-errors predicted by classifier (red). (b) Images from KITTI dataset are shown with detections in left and right cameras (cyan), disparity map, shifted detections (yellow), errors predicted by classifier (green) and non-errors predicted by classifier (red). [Best viewed in color]}
\end{figure*}

In this section, we present the technical details of our proposed system.
We study two distinct cues to identify errors - \emph{temporal} and \emph{stereo} inconsistencies - that are readily available in the AV object detection domain. 
\textbf{Temporal:} CNN based object detectors often fail to consistently detect objects in subsequent frames even with little motion. But region 
based trackers (e.g., ~\cite{kalal2012tracking}) are often able to reliably track a patch across frames even with significant occlusions or lighting variations.  
This allows us to use a multi-object tracker to estimate the location of missed objects in current frame using detections in previous frames (see Fig.~\ref{fig:pipeline_temporal}). 
\textbf{Stereo:} Though a pair of stereo images appear visually similar to one another, particularly when considering a small
baseline setup, object detectors often fail to consistently detect objects in both images (see Fig.~\ref{fig:pipeline_stereo}).
We use the diparity map to transfer detections from one image to another, allowing us to identify inconsistencies.
Note that both computing disparity maps and tracking a region between a pair of images does not require
any high level semantic information. Consequently, these operations are robust to small changes between the pair of images.
The following sections explain our system in detail. 
We begin by briefly defining our problem and then present our approaches to utilizing temporal and stereo cues.

\subsection{Problem Definition}
Consider a single frame object detector $\mathcal{D}$ that is trained on a large labeled dataset $\mathcal{L}$. Additionally, there is an unlabeled dataset $\mathcal{U}$ that is collected by continuously driving AVs equipped with cameras. In particular, we assume $\mathcal{U} = \{ I_j, j = 1,2,\ldots \}$ has images in sequence captured at a sufficiently high framerate to allow for tracking and each $I_j$ has a corresponding image $I_j'$ captured from the stereo camera. We employ $O_j = \{o_k^j, k=1,2,\ldots\}$ to denote the set of objects present in $I_j$ and $\hat{O}_j = \{\hat{o}_l^j, l=1,2,\ldots\}$ to denote objects detected by $\mathcal{D}$ on $I_j$ and similarly $\hat{O}_j'$ for detections on $I_j'$. The objective of this work is to reliably identify errors $E_j = \{e_i^j,i=1,2,\ldots\}$ made by $\mathcal{D}$ on $I_j \in \mathcal{U}$.
Recall that an error could either be a detection of a non-object (false positive) or a missed object (false negative). We consider only false negatives in this paper. There are promising approaches to false positive detection in the literature~\cite{barnes2017find} using multiple sensing modalities and free space detection. However, little work has gone into automated false negative detection and in the autonomous vehicle context false positives are more easily flagged by test drivers as they often cause the vehicle to stop or swerve for phantom objects. Additionally, for current state-of-the-art detectors, the number of false negatives are much greater than false positives\footnote{From the precision-recall curve on KITTI, note that detectors achieve $100\%$ precision for majority of recall values.}. 
Therefore, we focus this paper on identifying false negatives and henceforth use the term errors to refer to only false negatives of $\mathcal{D}$ (i.e) $E_j = O_j-\hat{O}_j$. Future work will address the integration of false positive results with the proposed approach.

\subsection{Temporal Cues}
\label{subsec:temporal}

In this subsection, we detail our approach for using an off-the-shelf multi-object tracker $\mathcal{T}$ trained on $\mathcal{L}$ to determine $E_j$. In detection based trackers, we have $\hat{T}_j = \mathcal{T}(I_j,\hat{O}_j, \hat{T}_{j-1})$ where $\hat{T}_j = \{\hat{t}_m^j, m=1,2,\ldots\}$ is the set of tracklets maintained by $\mathcal{T}$ for frame $j$.
Given $\hat{O}_j$ and $\hat{T}_j$, we define a set of hypothesis $H_j = \hat{T}_j - \hat{O}_j = \{h_n^j, n=1,2,\ldots\}$. As objects are commonly denoted by their bounding boxes, this set differencing operation can be implemented by performing bipartite matching between $\hat{T}_j$ and $\hat{O}_j$ using the Hungarian algorithm with the cost as $1-\mathtt{overlap}(\hat{t}_m^j,\hat{o}_l^j)$ where Intersection-over-Union (IoU) of the bounding boxes is used to compute the overlap. Here, we enforce a minimum of $50\%$ overlap for each match. Consequently, each $h_n^j$ corresponds to an unmatched tracklet. We propose that the task of determining $E_j$ can be approximated by a binary classification problem of determining if $h_n^j \in E_j$ or not. Tracker failures, which can be common, lead to false hypotheses which can appear as duplicate tracks or loose bounding boxes. This necessitates an approach to disambiguate these cases in the error detection process. 

\begin{figure}[!th]
    \centering
    \includegraphics[width=\linewidth]{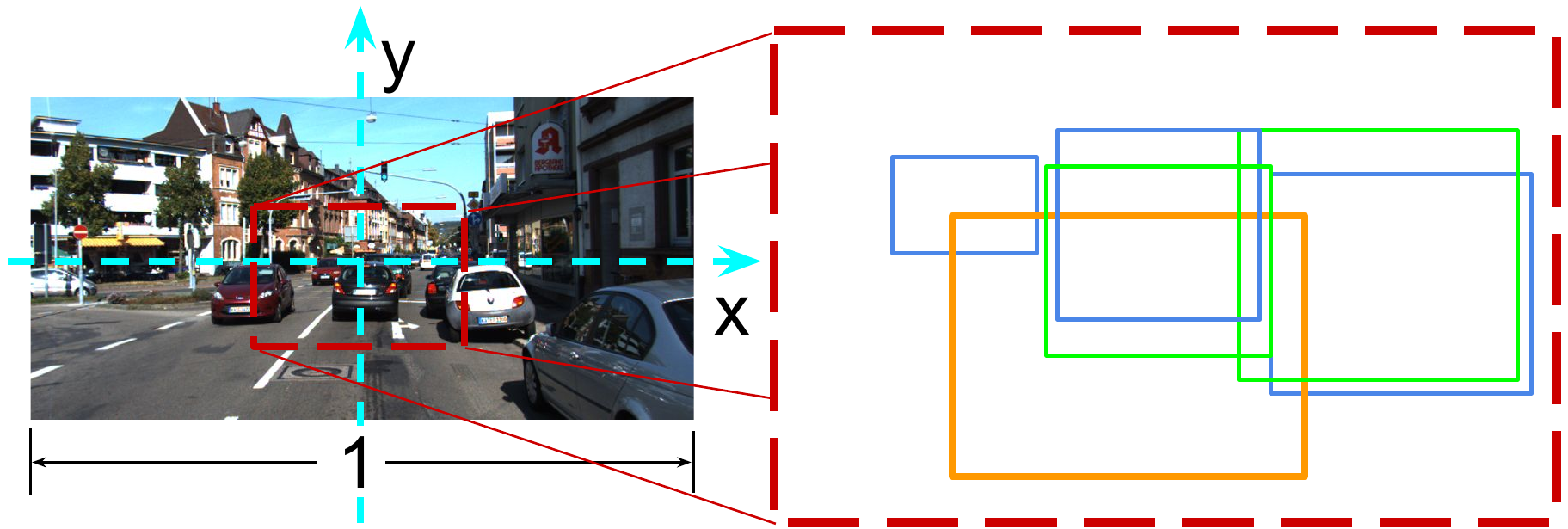}
    \caption{The proposed features describe the scenario around the hypothesis. The image is shown in the normalized coordinate space on the left and a region around the hypothesis is enlarged and shown on the right. The hypothesis in consideration is shown in orange, the overlapping detections in blue and the green boxes denote overlapping tracks (or shifted detections). See Section~\ref{subsec:feature_description} for detailed description.}
    \label{fig:features}
\end{figure}

\begin{table}[ht]
\centering
\begin{tabular}{|m{6em}|m{6cm}|}
\hline
\textbf{Name} & \textbf{Description} \\
\hline
$x,y$ & position of hypothesis in normalized coordinate space \\
\hline
$w,h$ & width and height of hypothesis in normalized scale \\
\hline
$r$ & confidence of hypothesis \\
\hline
$det\_cnt$ & number of detections overlapping the hypothesis \\
\hline
$med\_det\_ov$ & median overlap of above \\
\hline
$med\_det\_cnf$ & median confidence of above \\
\hline
$hyp\_cnt$ & number of overlapping $t_m^j$ or $\overleftarrow{o}_l^j$ \\
\hline
$med\_hyp\_ov$ & median overlap of above \\
\hline
$med\_hyp\_cnf$ & median confidence of above \\
\hline
$n$ & length of track (only for temporal cue) \\
\hline
\end{tabular}
\caption{The proposed set of features for the binary classification problem}
\label{tbl:features}
\end{table}

\subsection{Stereo Cues}
\label{subsec:stereo}

In this subsection, we detail our approach for using a disparity map generator $\mathcal{S}$, which can be a deterministic algorithm such as block matching or a learning based algorithm \cite{zbontar2016stereo} trained on $\mathcal{L}$, to determine $E_j$. For calibrated stereo cameras with known geometry,
we first find pixel-wise associations between the corresponding images using disparity map generator $\mathcal{S}$ to compute $d_j = \mathcal{S}(I_j, I_j')$. 
This allows us to \emph{shift} the detections $\hat{O}_j'$ to $I_j$, denoted as $\overleftarrow{O}_j$. In particular, we use the median disparity of the region covered by the bounding box which is robust to noise in disparity estimates. We now define the set of hypothesis for stereo cue as $H_j = \overleftarrow{O}_j - \hat{O}_j$ and follow the same procedure as in Section~\ref{subsec:temporal} wherein $\hat{T}_j$ is replaced with $\overleftarrow{O}_j$. However, the challenges posed by the stereo cue differ from the temporal cue. In particular, large errors in disparity estimates especially for small objects are a challenge. Moreover, false positives of the object detector could also be inconsistent between stereo views. Also, objects in the right image could be completely occluded in the left image. 

\subsection{Features for Classification problem}
\label{subsec:feature_description}
 In the previous subsections, we presented our approach to using stereo and temporal cues to compute hypothesis for errors $E_j$. Although the challenges for the classifier are different for the two cues, we propose a common set of $11$ features for each $h_n^j$ described in Table.~\ref{tbl:features}. Additionally, we also included the length of the track for the temporal cue but we will shown later in Section~\ref{sec:synthetic_data} that the classifier gives less importance to this feature. Fig.~\ref{fig:features} shows an illustration of the proposed features. The normalized coordinate space with the origin at the center of projection allows the features to be consistent for varying image dimensions and aspect ratios while preserving the location information from a canonical viewpoint in AVs. The power of such an approach is demonstrated in our experiments by the efficacy of these features across data sets with different cameras and image resolutions. In the example scene, there are three detections and two other tracks (or shifted detections) overlapping the selected hypothesis (orange) to various degrees. We use the count, median overlap and median confidence of the overlapping detections as well as overlapping tracks (or shifted detections) as features. Using these proposed features, an off-the-shelf binary classification algorithm (in this case a Random Forest Classifier~\cite{breiman2001random}), trained separately for temporal and stereo cues, can be used to identify $E_j$. Training details will be presented in Section~\ref{sec:synthetic_data}.


\section{Experiments}
\label{sec:result}

In this section, we present experimental results using our proposed system. In order to extensively evaluate the temporal and stereo cues, we require a large dataset consisting of \emph{sequences} of \emph{stereo} images with labels in addition to a labeled dataset for training the object detector, preferably from the same domain. Therefore we use the Sim200k dataset from \cite{johnson2017driving} for training the object detector and additionally generate a large tracking dataset from the same domain and call it the GTA dataset. Furthermore, we rendered \emph{stereo images} by using the depth buffer information from the game engine and performed in-painting \cite{bertalmio2001navier} to fill holes in these images. 
Note that the GTA dataset has no overlap with the Sim200k dataset and can be considered as newly collected data from a fleet of deployed test vehicles. The 104 sequences in GTA dataset is split into GTA12 and GTA92 containing 12 and 92 sequences respectively. In the following subsections, we use three state-of-the-art detectors, namely SSD~\cite{liu2016ssd}, Faster R-CNN~\cite{NIPS2015_5638} and RRC~\cite{ren2017accurate}\footnote{We used only a 10k subset for RRC as training is very slow} object detectors trained on the Sim200k dataset. We present all the analysis for the \texttt{Car} object category but the system can be trivially extended to other object categories as well.

Table~\ref{tbl:summary} provides a summary of the quantitative results of our experiments. In the following subsections, we discuss in detail about the different experiments that make up this table. We have included a supplementary MP4 format video which demonstrates our system in both synthetic and real datasets. 

\begin{table}[th]
\centering
\setlength\extrarowheight{2pt}
\begin{tabular}{|c|c|c|c|c|c|}
\hline
\multirow{2}{*}{\textbf{Cue Type}} & \multirow{2}{*}{\textbf{Detector}} & \multicolumn{2}{c|}{\textbf{GTA92}} & \multicolumn{2}{c|}{\textbf{KITTI}} \\[2pt] \cline{3-6}
 & & \textbf{Naive} & \textbf{Classifier} & \textbf{Naive} & \textbf{Classifier} \\[2pt]
\hline
\multirow{3}{*}{Temporal} & SSD & $0.44$ & $0.77$ & $0.65$ & $0.87$ \\
 & RCNN & $0.50$ & $0.85$ & $0.71$ & $0.92$ \\
 & RRC & $\textbf{0.67}$ & $\textbf{0.86}$ & $\textbf{0.87}$ & $\textbf{0.94}$ \\
\hline
\multirow{3}{*}{Stereo} & SSD &  $0.47$ & $0.77$ & $0.59$ & $0.77$ \\
 & RCNN & $0.50$ & $0.81$ & $0.54$ & $0.82$ \\
 & RRC & $\textbf{0.81}$ & $\textbf{0.93}$ & $\textbf{0.81}$ & $\textbf{0.88}$ \\
\hline
\end{tabular}
\caption{Average Precision scores for the binary classification task using Stereo and Temporal cues. Naive approach corresponds to flagging all hypotheses as errors}
\label{tbl:summary}
\end{table}

\subsection{Baseline Experiments}
\label{sec:baseline}

The GTA dataset consists labels even for small objects that are typically far away from the ego-vehicle. Therefore, we ignore all objects that have a height less than $25\mathtt{px}$ (same as the KITTI benchmark). The average precision of the detectors ($>50\%$ overlap) on GTA92 are: SSD (0.50), RCNN (0.66) and RRC (0.80). First, we need to select a confidence threshold for each detector. Only detections above this threshold are tracked (temporal) or shifted using disparity (stereo). For SSD and RRC, we show good performance even for a low threshold of $0.5$. However RCNN makes many false positives with high confidence and therefore we use a high threshold of $0.98$. 

It is important to note, to the best of our knowledge, we have found no automated error detection scheme for AV object detectors in the literature so comparison to other approaches is difficult. 
However, we provide a comparison to a naive approach that uses the two proposed cues directly without the additional meta-feature classification of potential errors i.e. all hypotheses would be flagged as errors. 
Although both the hypothesis generation and subsequent binary classification are contributions of our work.

\subsection{Quantitative analysis on GTA Dataset}
\label{sec:synthetic_data}
In this section, we present quantitative results for the performance of temporal and stereo cue in finding object detector errors. We repeat the experiment for each of the three detectors separately.

\subsubsection{Temporal Cue} 
Given the detections on the GTA dataset, we train the MDP~\cite{Xiang_2015_ICCV} Multi-object Tracker on GTA12. With the trained tracker, we generate hypotheses for all the 104 sequences in GTA dataset using the approach in Section~\ref{subsec:temporal}. For each hypothesis, we compute the proposed $12$ features along with the ground truth label - valid/invalid error. Here, we can use the labels from the tracking dataset to compute labels for the binary classification task. The hypotheses from the GTA12 are used to train a random forest classifier. The remaining GTA92 dataset is used to evaluate both the naive approach as well as the trained classifier.  

\begin{figure}[t]
    \centering
    \begin{subfigure}[t]{0.4\linewidth}
        \includegraphics[width=\linewidth]{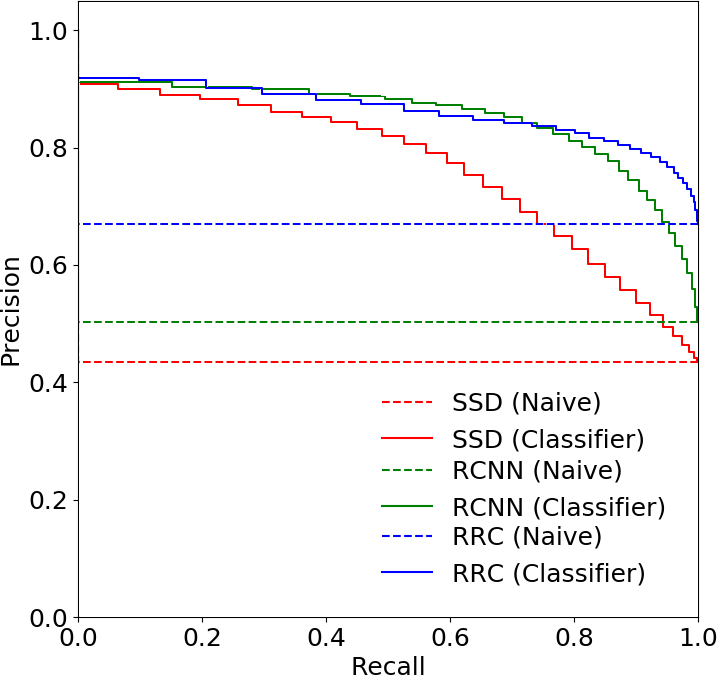}
        \caption{Temporal on GTA92}
        \label{fig:pr_curve_temporal}
    \end{subfigure}~
    \begin{subfigure}[t]{0.4\linewidth}
        \includegraphics[width=\linewidth]{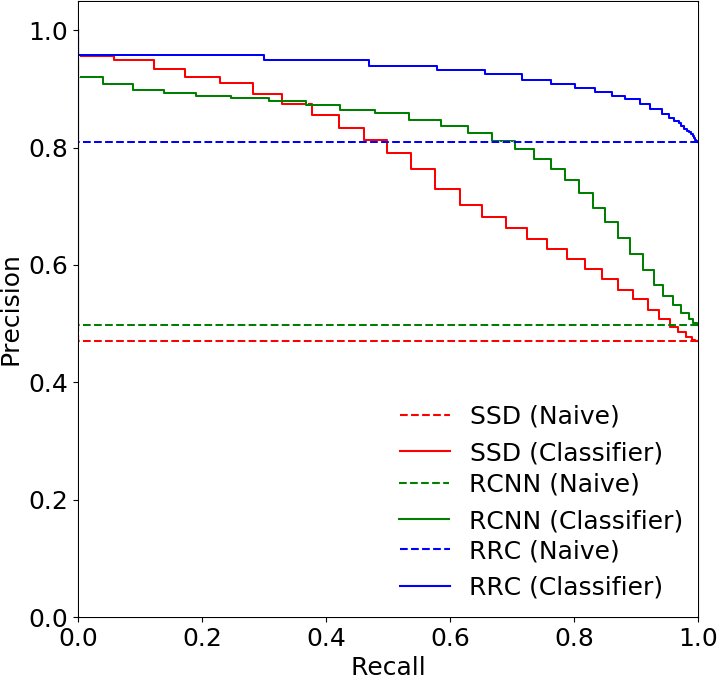}
        \caption{Stereo on GTA92}
        \label{fig:pr_curve_stereo}
    \end{subfigure}
    \begin{subfigure}[t]{0.4\linewidth}
        \includegraphics[width=\textwidth]{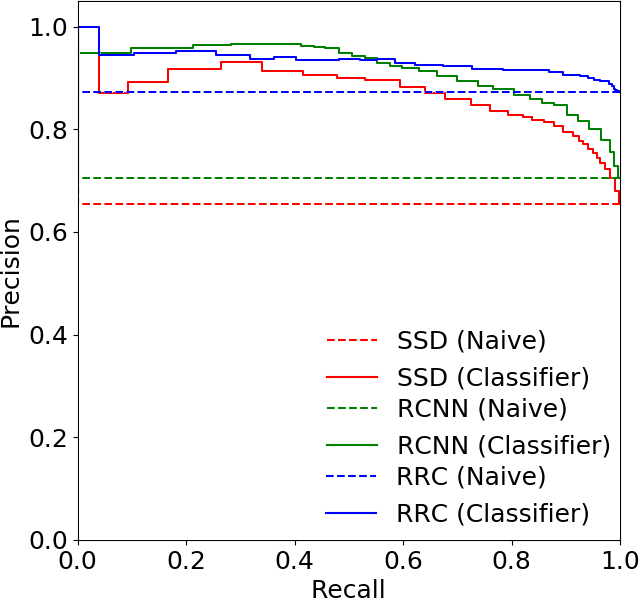}
        \caption{Temporal on KITTI}
        \label{fig:pr_curve_temporal_kitti}
    \end{subfigure}~
    \begin{subfigure}[t]{0.4\linewidth}
        \includegraphics[width=\textwidth]{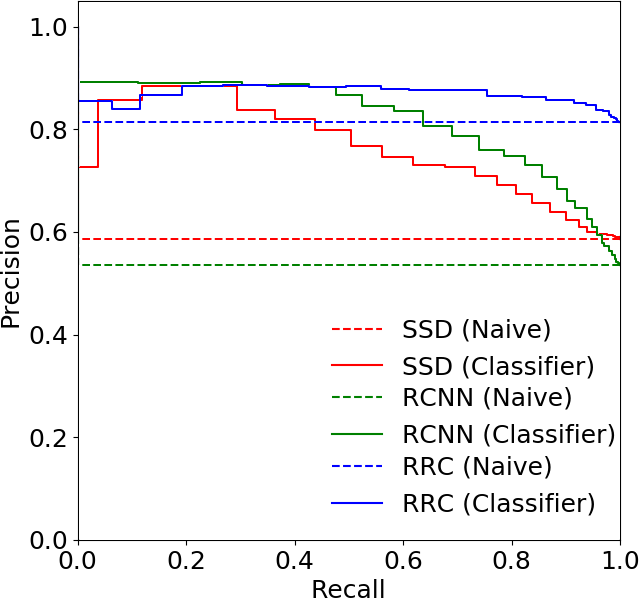}
        \caption{Stereo on KITTI}
        \label{fig:pr_curve_stereo_kitti}
    \end{subfigure}
    \caption{Precision-Recall curves for Random Forest Classifiers with 30 estimators. }
    \label{fig:random_forest_classifier}
\end{figure}

Fig.~\ref{fig:pr_curve_temporal} shows the precision recall curves for the classifiers corresponding to the three detectors. As the accuracy of the base object detector improves, the accuracy of the classifier improves and also the gap between the naive approach and the classifier decreases (refer Table~\ref{tbl:summary}). 
Although the feature importances vary for each detector, the dominant features are the median overlap and confidence of the overlapping tracks followed by the position and size of the hypothesis. Interestingly, the length of the track and its associated confidence are not the dominant features in this binary classification task.

\subsubsection{Stereo Cue} For the stereo cue, we use the SemiGlobal Block Matching algorithm~\cite{hirschmuller2008stereo} implemented in OpenCV to compute the disparity maps.
We generate hypotheses for all the 104 sequences and compute the proposed $11$ features for each stereo hypothesis along with ground truth labels. We train a random forest classifier using hypotheses from GTA12 and evaluate the trained classifier as well as the naive approach on GTA92.
Fig.~\ref{fig:pr_curve_stereo} shows the precision recall curves for the classifiers corresponding to each detector. The accuracy of both the naive approach and classifier improves with accuracy of the detector (refer Table~\ref{tbl:summary}).
The dominant features for the stereo cue are the overlaps, position and size of the hypotheses.

\subsubsection{Stereo and Temporal cue}
The striking similarity in the behavior of stereo and temporal cues led us to compare the errors detected by each cue. 
We concatenate the identified errors from each cue and apply non-maximum suppression with a minimum overlap of $0.7$ IoU. For RRC detector, while the temporal and stereo cue predicted $13,205$ and $8,572$ errors respectively, of those $10,568$ and $5,935$ were unique to each cue respectively showing us the complementary nature of the two cues to one another. We observed a similar trend for other two detectors as well. 

\subsection{Real world Labeled Datasets}
\label{sec:real_data}

In this experiment, we use the system (detector, tracker and classifier) trained on the GTA12 dataset and evaluate the performance on the KITTI tracking dataset. We use the ground truth labels from the KITTI dataset only to evaluate the accuracy of the classifier and never for training. 
The underlying object detectors (simulation only trained) achieve AP scores of $0.45$ (SSD), $0.54$ (RCNN) and $0.68$ (RRC) on the KITTI dataset where $>70\%$ overlap is required for correct detection. As the GTA dataset consider large vehicles like trucks as the \texttt{Car} category, we collapse \texttt{Truck}, \texttt{Van} and \texttt{Car} in the KITTI labels as well. 

As seen in Table~\ref{tbl:summary}, both the naive approach and the trained classifier perform well on real world data as well. Fig.~\ref{fig:pr_curve_temporal_kitti} and Fig.~\ref{fig:pr_curve_stereo_kitti} show the corresponding precision recall curves for the classifiers. Interestingly, the performance of the system is better in KITTI dataset than the GTA dataset. This is likely because the challenging GTA dataset contains many small objects with heavy occlusions and truncations as opposed to KITTI dataset where those objects would be ignored during evaluation. Similar to the comparison between temporal and stereo cues in the GTA dataset, we computed the overlap between errors identified by the two cues. For the RRC detector, while the temporal and stereo cues predicted $1,335$ and $1,509$ errors respectively, only $493$ were in the intersection of both sets, again suggesting the complementary nature of both cues. We observed a similar trend for the other two detectors as well. 

In order to validate that the identified errors improve object detection, we compute the F1-score at the chosen confidence thresholds of the detectors with and without the stereo and temporal errors identified from our system (refer Table~\ref{tbl:f1_scores}). As our approach primarily improves the recall of the detector, we chose the F1-score. Since region based tracking and disparity based shifting of bounding boxes could introduce localization errors, we relax the overlap requirement to $>50\%$ for this evaluation. 

\begin{table}[th]
\centering
\setlength\extrarowheight{2pt}
\begin{tabular}{|c|c|c|c|}
\hline
\textbf{Detections} & \textbf{SSD} & \textbf{RCNN} & \textbf{RRC} \\
\hline
w/o our system & $69.63\%$ & $80.15\%$ & $83.95\%$ \\
w/ our system & $73.85\%$ & $81.02\%$ & $87.52\%$ \\
\hline
Improvement & $4.22\%$ & $0.87\%$ & $3.57\%$ \\
\hline
\end{tabular}
\caption{F1 scores of the detectors with and without the error correction}
\label{tbl:f1_scores}
\end{table}

\subsection{Real world Unlabeled datasets}
\label{sec:unlabeled_data}

We now consider the three detectors trained on the KITTI object recognition dataset and use our proposed system on the unlabeled KITTI tracking test set and the Oxford RobotCar~\cite{maddern20171} dataset.\footnote{The sequences are indexed in RobotCar dataset using the date and time when the data was collected. We used the following sequences: Overcast (2015-08-13-16-02-58, 2015-10-30-13-52-14, 2015-05-19-14-06-38), Sunny (2015-02-17-14-42-12, 2015-03-24-13-47-33, 2015-08-12-15-04-18)}
For each detector, we independently train both the MDP tracker and our classifiers on the KITTI tracking dataset. Some examples of the found errors are shown in Table~\ref{tbl:mistakes_mosaic}.

We consider two weather conditions from the RobotCar dataset - Overcast and Sunny. All three detectors suffer from domain transfer and find fewer objects in the RobotCar dataset. However, our system finds errors in both weather conditions. For the RRC detector, our system predicted $6,272$ errors in overcast images and $5,798$ errors in sunny images given about same number of images of each kind.

\begin{table*}[!t]
\setlength\tabcolsep{0pt}
\centering
\begin{tabular}{m{1.5cm} c c c c c}
\multicolumn{6}{c}{\textbf{Temporal Cue - RobotCar}} \\
\hline
\textbf{Detector} & \textbf{Shadow} & \textbf{Saturated} & \textbf{Occluded} & \textbf{Visible} & \textbf{False Positive} \\
\hline
SSD \cite{liu2016ssd} & \raisebox{-0.5\height}{\includegraphics[width=0.16\textwidth]{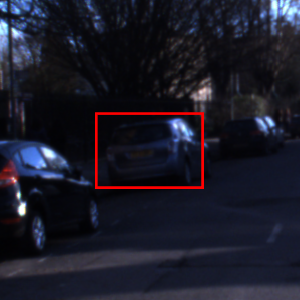}} & \raisebox{-0.5\height}{\includegraphics[width=0.16\textwidth]{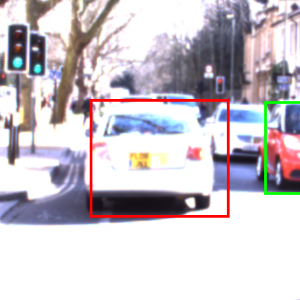}} & \raisebox{-0.5\height}{\includegraphics[width=0.16\textwidth]{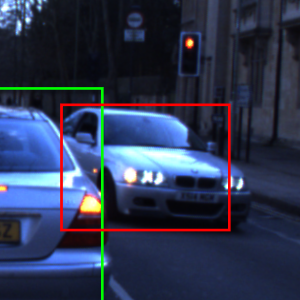}} & \raisebox{-0.5\height}{\includegraphics[width=0.16\textwidth]{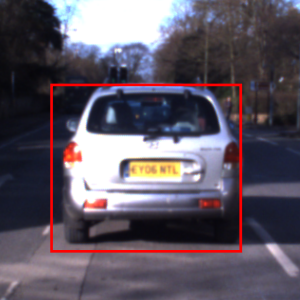}} & \raisebox{-0.5\height}{\includegraphics[width=0.16\textwidth]{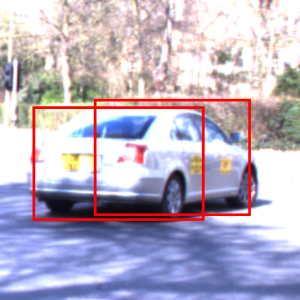}} \\ 
Faster RCNN \cite{NIPS2015_5638} & \raisebox{-0.5\height}{\includegraphics[width=0.16\textwidth]{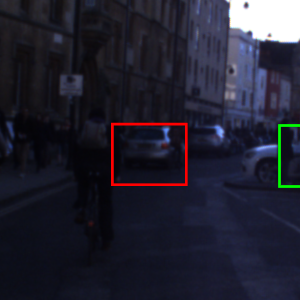}} & \raisebox{-0.5\height}{\includegraphics[width=0.16\textwidth]{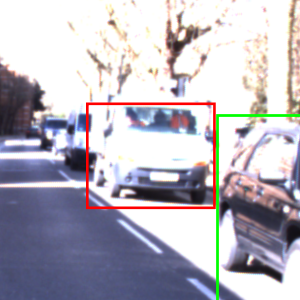}} & \raisebox{-0.5\height}{\includegraphics[width=0.16\textwidth]{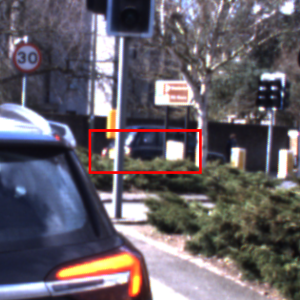}} & \raisebox{-0.5\height}{\includegraphics[width=0.16\textwidth]{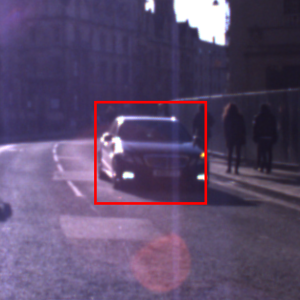}} & \raisebox{-0.5\height}{\includegraphics[width=0.16\textwidth]{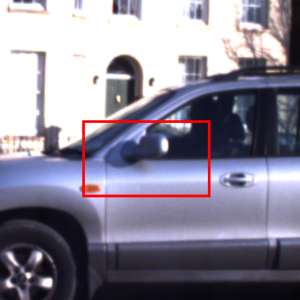}} \\ 
RRC \cite{ren2017accurate} & \raisebox{-0.5\height}{\includegraphics[width=0.16\textwidth]{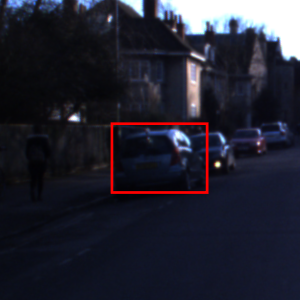}} & \raisebox{-0.5\height}{\includegraphics[width=0.16\textwidth]{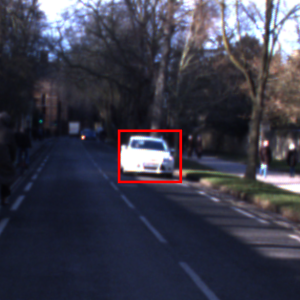}} & \raisebox{-0.5\height}{\includegraphics[width=0.16\textwidth]{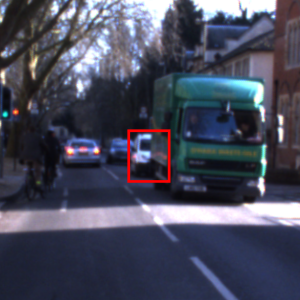}} & \raisebox{-0.5\height}{\includegraphics[width=0.16\textwidth]{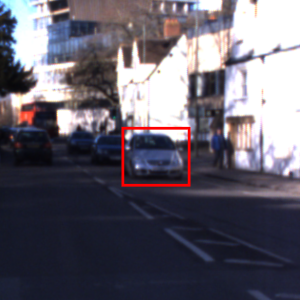}} & \raisebox{-0.5\height}{\includegraphics[width=0.16\textwidth]{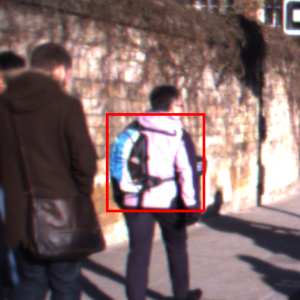}} \\ 
\multicolumn{6}{c}{\textbf{Stereo Cue - KITTI}} \\
\hline
\textbf{Detector} & \textbf{Shadow} & \textbf{Saturated} & \textbf{Occluded} & \textbf{Visible} & \textbf{False Positive} \\
\hline
SSD \cite{liu2016ssd} & \raisebox{-0.5\height}{\includegraphics[width=0.16\textwidth]{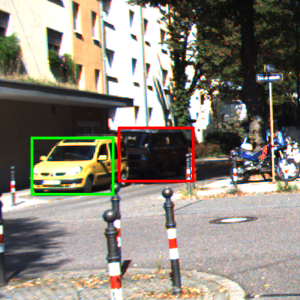}} & \raisebox{-0.5\height}{\includegraphics[width=0.16\textwidth]{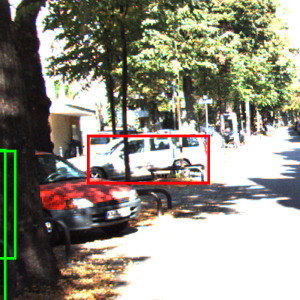}} & \raisebox{-0.5\height}{\includegraphics[width=0.16\textwidth]{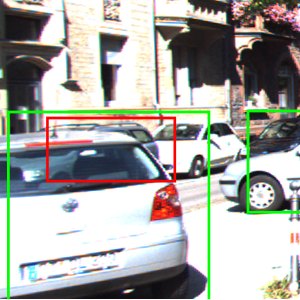}} & \raisebox{-0.5\height}{\includegraphics[width=0.16\textwidth]{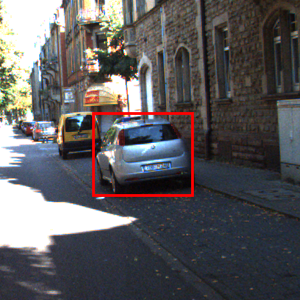}} & \raisebox{-0.5\height}{\includegraphics[width=0.16\textwidth]{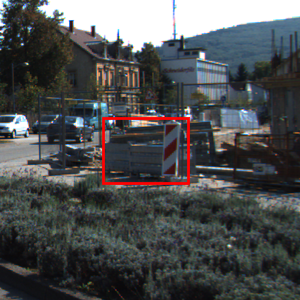}} \\
Faster RCNN \cite{NIPS2015_5638} & \raisebox{-0.5\height}{\includegraphics[width=0.16\textwidth]{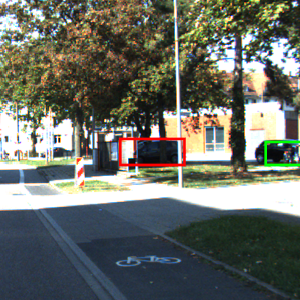}} & \raisebox{-0.5\height}{\includegraphics[width=0.16\textwidth]{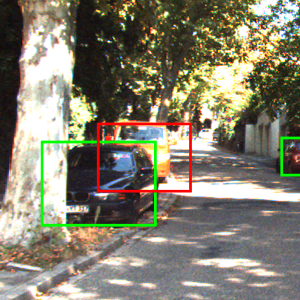}} & \raisebox{-0.5\height}{\includegraphics[width=0.16\textwidth]{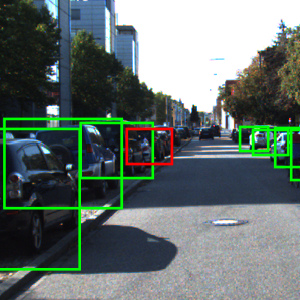}} & \raisebox{-0.5\height}{\includegraphics[width=0.16\textwidth]{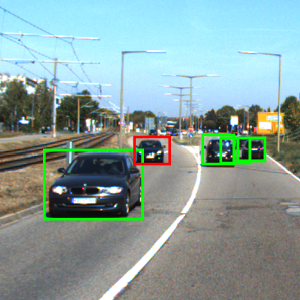}} & \raisebox{-0.5\height}{\includegraphics[width=0.16\textwidth]{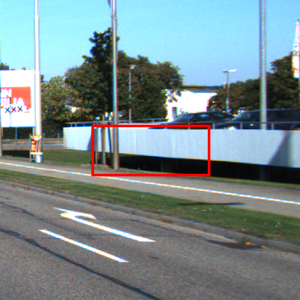}} \\
RRC \cite{ren2017accurate} & \raisebox{-0.5\height}{\includegraphics[width=0.16\textwidth]{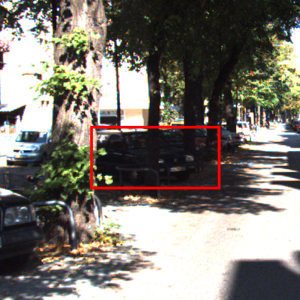}} & \raisebox{-0.5\height}{\includegraphics[width=0.16\textwidth]{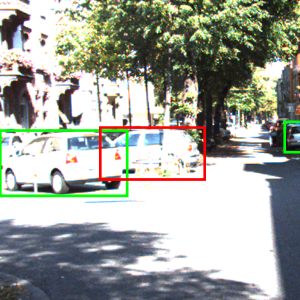}} & \raisebox{-0.5\height}{\includegraphics[width=0.16\textwidth]{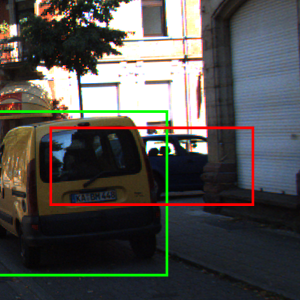}} & \raisebox{-0.5\height}{\includegraphics[width=0.16\textwidth]{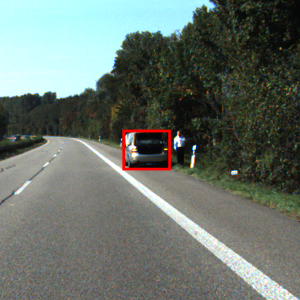}} & \raisebox{-0.5\height}{\includegraphics[width=0.16\textwidth]{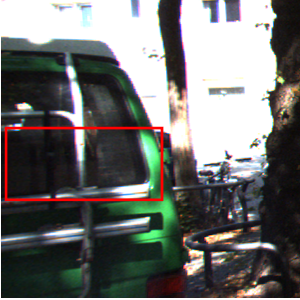}} \\
\end{tabular}
\caption{Examples of identified errors for different detectors using temporal cue on RobotCar dataset and stereo cue from the KITTI dataset. These images have been manually selected and each image is a $300\times 300$ crop centered around the missed car (shown in red). The green boxes are detections from the object detector. The size of the crops was fixed in order to show the different scales at which mistakes were identified. [Best viewed in color]}
\label{tbl:mistakes_mosaic}
\end{table*}

\subsection{Correlations in detected errors to Geo-location}
\label{sec:spatial}

\begin{figure}[th]
    \centering
    \begin{subfigure}[t]{0.4\linewidth}
        \includegraphics[width=\linewidth]{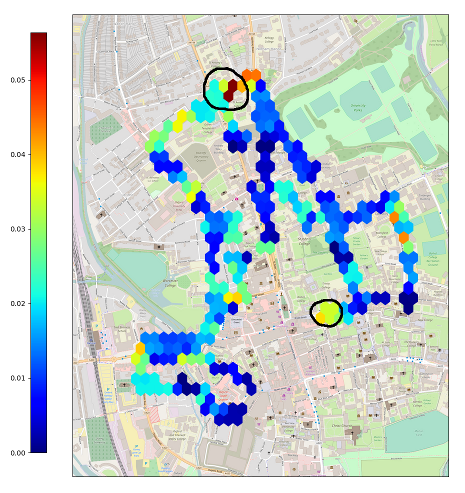}
        \caption{Oxford}
        \label{fig:oxford}
    \end{subfigure}
    \begin{subfigure}[t]{\linewidth}
        \includegraphics[width=\linewidth]{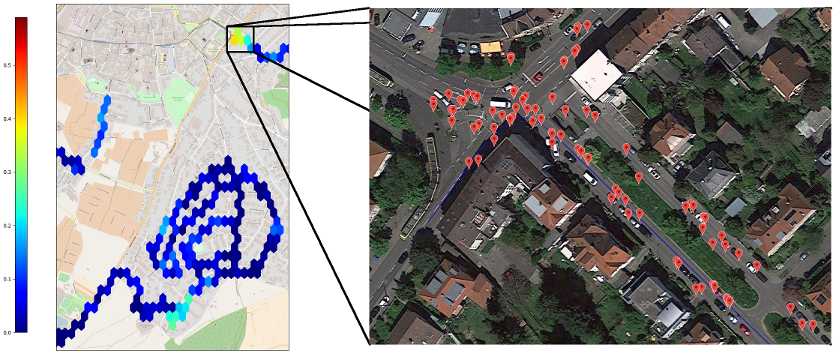}
        \caption{KITTI}
        \label{fig:kitti}
    \end{subfigure}
    \caption{Correlations of errors detected by our system to their geo-locations. (a) black circles highlight certain intersections in Oxford that prove more difficult than others and warrant increased testing. (b) displays the spatial fidelity with which missed detections can be localized in the KITTI dataset with the incorporation of range data from the perception system. The red flags in the call out correspond to precise metric locations where predicted missed cars were present.}
    \label{fig:spatial_errors_map}
\end{figure}

The proposed system can be used to detect systematic errors of an object detector with respect to geo-location. Here we display results that highlight the utility of such a system for testing and development purposes.
We use the sensor data in KITTI tracking and RobotCar datasets to localize mistakes found using the temporal cues. By binning the vehicle's pose estimates, we can then calculate the average number of errors per frame in various map regions. Heatmaps of the mistakes found appear (shown in Fig.~\ref{fig:oxford}) to highlight that several intersections have an error rate of roughly twice that of other seemingly similar intersections along the vehicle's route.
Using the KITTI data we can harness additional range data information, to localize errors to precise spatial coordinates enabling the identification of each missed car to the lane and trajectory through the intersection (see Fig.~\ref{fig:kitti}). 

\section{Discussion}
\label{sec:discussion}
The experimental results demonstrate the efficacy of our method in detecting real
failures of state-of-the-art object detectors on unlabeled data. 
There are some significant and immediate advantages to our method that have application for currently deployed AV fleets. Our method can be directly integrated with any object detector, 
multi-object tracker and stereo disparity computation method. So any deployed AV system that relies on that underlying technology (which is almost all that have and use cameras) can benefit from this technique for data mining. As the proposed features and the classifier are lightweight ($5 \mathtt{ms}/$frame with unoptimized code), the system can be integrated for real-time usage. We posit that our system could greatly increase safety and efficiency of both testing and training AVs with little overhead or additional infrastructure. 

There are some limitations to our system as well. In particular, if the detector never detects an object, then our system will not be able to detect those errors. 
Our system is also limited by the challenges of multi-object tracking such as crowded scenes with interleaving motion and heavily
occluded objects\cite{luo2014multiple}. Inconsistent false positives of object detectors would also cause false positives for our system.

The ability to correlate errors to geo-locations offers the ability for AV testing to be done in a more focused manner when it comes to understanding object detection failure. Correlating this information with time of day, direction of travel, or other physical properties of the area of failure can have immense benefit in understanding when your system is failing. Examples such as tree-line streets, driving into the sun, or certain angles-of-attack with respect to oncoming traffic could all be identified automatically. Beyond this the capability of gathering massive datasets of failure cases from a fleet of AVs could transform the way manual labeling is done. Eliminating the need for hundreds of hours of human labor.


\section{Conclusion}
\label{sec:conclusion}
We presented a system for self-driving cars that enables checking for inconsistency using two distinct mechanisms: temporal and stereo cues. Our proposed system provides a means of identifying false negatives made by single frame object detectors on large unlabeled datasets. We propose that finding object detector errors can be posed as a binary classification problem. 
We use an off-the-shelf multi-object tracker to construct tracklets and each tracklet without an associated detection is used as a hypothesis. We use stereo disparity to shift detections from one camera view to the other and use the unassociated shifted detections as hypothesis for missed objects. We show that the naive approach of using inconsistencies is in itself a strong indicator of object detector errors as object detectors become more accurate. Additionally, we show that, using our proposed features, an off-the-shelf random forest classifier achieves an AP score of $0.93$ on the GTA dataset for RRC detector. Furthermore, we showed that our system (detector, tracker and classifier) trained only on synthetic data can find errors made in the KITTI dataset with an AP score of $0.94$ for RRC detector. This offers the promising of bootstrapping dataset labels for new domains through a process of synthetic training and failure mining in real data. Through extensive experiments we have shown that even the state-of-the-art object detectors make systematic errors and we can reliably localize these in a global reference frame.

Naturally, the next step is to make object detectors learn from these identified mistakes. This is a deceptively hard task for CNN based object detectors. In a supervised learning setting, the images are assumed to be exhaustively labeled. Any region in the image without a label is assumed to be a negative sample while the labels themselves are considered positive samples with tight bounding boxes. While our method reliably detects false negatives, it does not always detect all mistakes in the image.
We plan to address how best to learn from this partial information in our subsequent research work. Additionally, we plan to incorporate free space computation from the path of the vehicle and from active sensor returns like LIDAR to identify false positives to further improve our assessment and understanding of modern object detectors at the fleet level.








\bibliographystyle{IEEEtran}
\bibliography{IEEEabrv,root}

\end{document}